\documentclass{article}

 \usepackage[preprint]{neurips_2025}

\usepackage[utf8]{inputenc} 
\usepackage[T1]{fontenc}    
\usepackage{hyperref}       
\usepackage{url}            
\usepackage{booktabs}       
\usepackage{amsfonts}       
\usepackage{nicefrac}       
\usepackage{microtype}      
\usepackage{xcolor}         
\usepackage{amsmath}
\usepackage{soul}
\usepackage[style=ieee]{biblatex}
\usepackage{graphicx}
\usepackage{siunitx}

\title{Ultra-Efficient Decoding for End-to-End Neural Compression and Reconstruction}

\author{%
  Ethan G.~Rogers, Cheng Wang \\
  Department of Electrical and Computer Engineering \\
  Iowa State University\\
  Ames, IA 50011 \\
  \texttt{kneehaw@iastate.edu, chengw@iastate.edu}\\
}

\begin{document}

\maketitle

\begin{abstract}

Image compression and reconstruction are crucial for various digital applications. While contemporary neural compression methods achieve impressive compression rates,
the adoption of such technology has been largely hindered by the complexity and large computational costs of the convolution-based decoders during data reconstruction. 
To address the decoder bottleneck in neural compression, we develop a new compression-reconstruction framework based on incorporating low-rank representation in an autoencoder with vector quantization. We demonstrated that performing a series of computationally efficient low-rank operations on the learned latent representation of images can efficiently reconstruct the data with high quality.
Our approach dramatically reduces the computational overhead in the decoding phase of neural compression/reconstruction, essentially eliminating the decoder compute bottleneck while maintaining high fidelity of image outputs.


\end{abstract}
\vspace{-15pt}
\section{Introduction}
\begin{figure}[h]
  \centering
  \includegraphics[width=.7\linewidth]{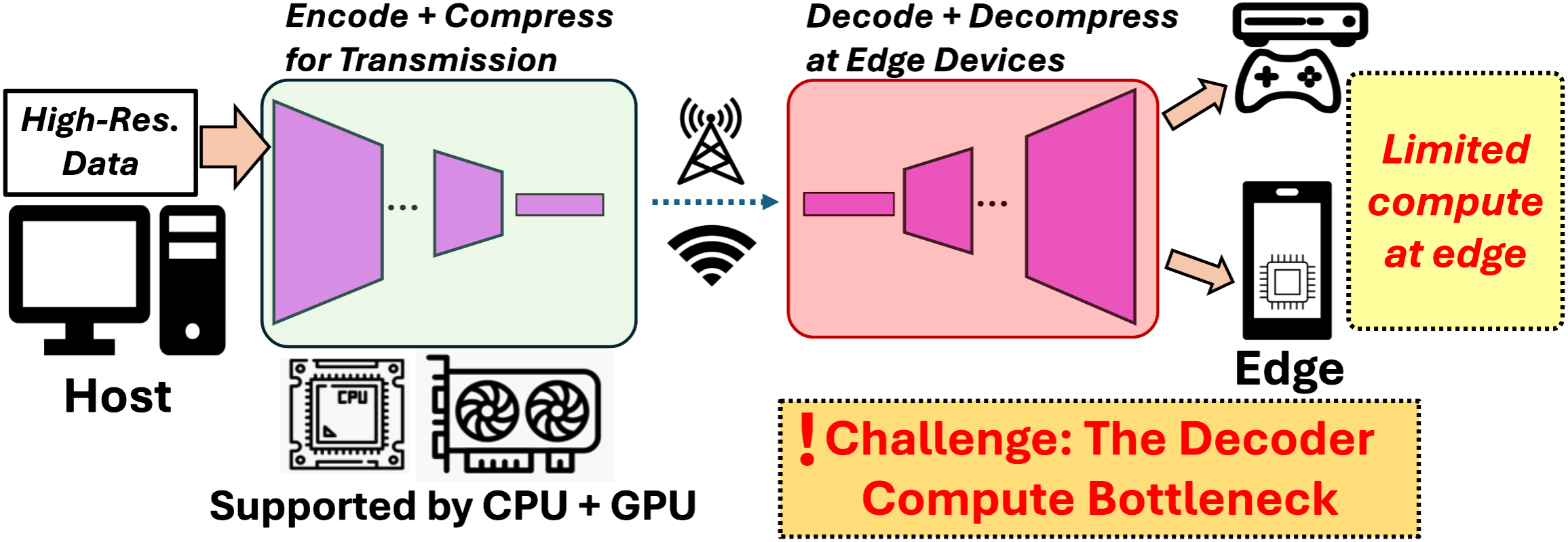}
  \vspace{-5pt}
  \caption{Data transmission and decode compute bottleneck.}
  \label{fig:motivation}
\end{figure}

\vspace{-8pt}
Image compression and reconstruction critically impact the quality and performance of many applications that are deployed to edge and mobile devices (such as streaming and gaming). Compared to the standard image compression codecs (such as JPEG), neural compression methods excel at learning the application-specific features adaptively and offering the desirable flexibility for further downstream tasks beyond data compression \cite{oord2017discrete}. 
As a result, neural compression may potentially lead to higher perceptual quality at the iso-compressed rate, and the representation learning from compression could be utilized for more sophisticated downstream tasks such as content generation. 
However, despite the promise of the AI-based compression, implementing such models encounters significant challenges, especially when they are deployed for edge applications. Specifically, most neural compression-reconstruction frameworks rely on consecutive convolutional blocks in both the encoder and decoder. While these compute-intensive convolutional blocks can be handled by graphic processing units (GPUs) during model development, such near-symmetric encoder-decoder neural architecture will impose a significant computational bottleneck when the decoder of these compression-reconstruction models needs to be computed on resource-constrained devices (Fig.\ref{fig:motivation}).   
Therefore, the performance and energy efficiency of running neural compression during reconstruction becomes extremely limited by the availability of hardware resources, preventing practical widespread use of these intelligent compression models.



In this work, we propose an end-to-end framework to address the decoder compute bottleneck. Low-rank representation is incorporated in an autoencoder with vector quantization to deliver extremely lightweight computation at the reconstruction step. The overall framework is also redesigned to accommodate the integration of low-rank vector representation. Our major contributions are the following:
\begin{itemize}
    \item We alleviate a significant portion of decoding compute by performing a learnable low-rank approximation of the output directly from an encoded latent representation. 
    We demonstrate that a modified transformer-based encoder with vector quantization and low-rank allows an ultra-lightweight decoding scheme to reach on-par output quality. 
    \item Application-specific parameters are introduced to allow for explicit trade-offs between output fidelity and decoder computational overhead. In addition to typical encoder hidden dimensions or codebook size, we also demonstrate explicitly that the reconstruction rank, iterations, as well as the size of patches in the transformer-based encoder proove effective knobs to adjust the compression-reconstruction model's level of granularity.
    
    \item We achieve a high degree of compression while maintaining quality, reaching over 21x reduction in image size at as low as $\num{3.6e-3}$ mean squared error (MSE). 
    Our approach shows that the aforementioned knobs combined can reach competitive compression capability while reducing the decoder computation by 10-100x when compared to state-of-the-art methods such as an adapted Google's Deepmind Sonnet VQVAE \cite{sonnet}.

\end{itemize}

\vspace{-15pt}

\section{Related Work}
\label{sec:codebook}
Vector quantization (VQ) has been a cornerstone of neural compression, mapping continuous data to discrete vectors from a shared codebook. Recent advancements have integrated this concept into deep learning architectures, most notably the Vector Quantized Variational Autoencoder (VQVAE) \cite{oord2017discrete} and the Vector Quantized Generative Adversarial Network (VQGAN) \cite{esser2021taming}. These models use a neural encoder to map input data into a continuous latent space that whose vectors are mapped to nearest neighbors from the codebook.
A decoder architecture then performs reconstruction on this quantized latent representation. This process converts the continuous-to-discrete mapping into a learnable end-to-end task, enabling powerful data compression and reconstruction. Some frameworks employing adjacent methods can see up to 240x compression while maintaining decent quality \cite{li2025onceforall}. Our work draws inspiration from this VQ paradigm but distinguishes itself by specifically addressing the decoder's computational cost. 

In approaches like Residual Vector Quantization (RVQ) the principle of iterative reconstruction with residuals is leveraged to produce greater fidelity. \cite{zeghidoursoundstream2021}. These works utilize a cascaded series of codebooks, each responsible for a different level of granularity in the latent VQ representation. We adapt this where instead of learning multiple codebooks for each residual layer, we use the output of a single codebook to parametrize an iterative series of low-rank matrix decompositions.

\begin{figure}
  \centering
  \includegraphics[width=.9\linewidth]{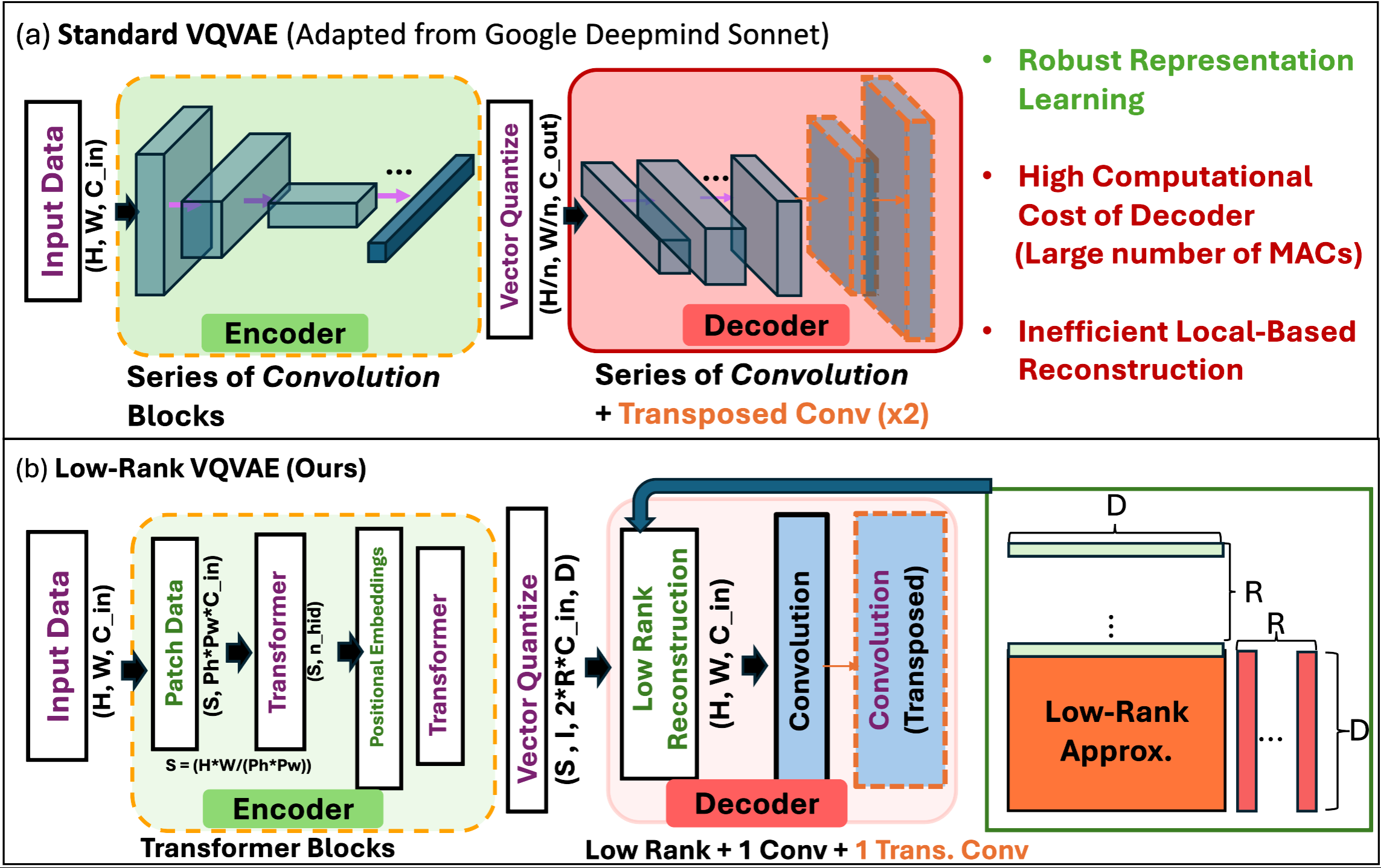}
  \caption{A comparison between (a) standard VQVAE architectures and (b) our low-rank VQVAE architecture. A low-rank operation is visualized in the bottom right.}
  \vspace{-16pt}
  \label{fig:framework}
\end{figure}

\section{Framework for Ultra-Efficient Reconstruction}

Singular-Value Decomposition (SVD) decomposes a matrix $A$ into three matrices, $\mathbf{U}\mathbf{\Sigma}\mathbf{V}^T$, where $\mathbf{U}$ and $\mathbf{V}$ are orthogonal matrices, and $\mathbf{\Sigma}$ is a diagonal matrix of singular values \cite{Beltrami1873}. A low-rank approximation of any matrix $\mathbf{A}$ can be obtained by retaining only the top $R$ singular values and their corresponding singular vectors, as shown by the Eckart-Young theorem \cite{Eckart_Young_1936}. This principal can be extended wherein a matrix is iteratively approximated by observing the residual error among a series of summed low-rank reconstructions. 

Given some initial target matrix $T_0\in\mathbb{R}^{m\times n\times C}$, where C is channels, the matrix can be reconstructed over iterations $i\in \{0,1, 2,\dots,I-1\}$:
\begin{equation}
    \label{eq:SVD}
    \mathbf{U}_{i}\mathbf{\Sigma}_{i}\mathbf{V}_{i}^T=\text{SVD}(\mathbf{T}_{i})
\end{equation}
\begin{equation}  
    \label{eq:SVDLR}
    \mathbf{T}_{i_{R}}=\mathbf{U}_{i_R}\mathbf{\Sigma}_{i_R}\mathbf{V}_{i_R}^T
\end{equation}
\begin{equation}
    \label{eq:residual}
    \mathbf{T}_{i+1}=\mathbf{T}_i-\mathbf{T}_{i_{R}}
\end{equation}

where $\mathbf{U}_{i}\in\mathbb{R}^{m\times m}, \mathbf{\Sigma}_{i}\in\mathbb{R}^{m\times n}, \mathbf{V}^T_{i}\in\mathbb{R}^{n\times n}$ and $\mathbf{U}_{i_R}\in\mathbb{R}^{m\times R}, \mathbf{\Sigma}_{i_R}\in\mathbb{R}^{R\times R}, \mathbf{V}^T_{i_R}\in\mathbb{R}^{R\times n}$, with $R$ corresponding to the target rank $R\in[1, \text{min}(m, n)]$, and \textbf{$\mathbf{A}_{i_R}$ }denoting the low rank version of any matrix \textbf{$\mathbf{A}$}. Following Eq. \ref{eq:residual}, 
once all $I$ low-rank approximations have been made, they produce the final reconstruction $\hat{\mathbf{T}}=\frac{1}{I}\sum_{i=0}^{I}\mathbf{T}_{i_{R}}$. The reconstruction can be evaluated by comparing $\hat{T}$ with $T_0$, where the element-averaged p-norm error is given by $\frac{1}{N}(\sum_{elems}(|\mathbf{T}_{0}-\hat{\mathbf{T}}|^{p}))^{\frac{1}{p}}$. 


The framework does not explicitly perform SVD. Instead, the encoder network learns to produce a compact latent representation of the required dimensions to perform low-rank reconstruction. These components come from a single vector codebook, allowing representation as just an index. The quantized latent vectors are used to generate the matrices $(\mathbf{U}_{i_R}\mathbf{\Sigma}_{i_R})$ and $(\mathbf{V}_{i_R})$ for each iteration. A high-quality latent representation is necessary for this application-specific task, hence our opting for a transformer over encoder frameworks. This design allows for a trainable framework with standard backpropagation techniques.


To retain fidelity for information-rich data, we introduce patching techniques similar to \cite{vit_dosovitskiy2021imageworth16x16words}. Then, reconstruction task is performed on smaller matrices, limiting the quantity of unique low-rank vectors needed to reproduce a variety of patches. Artifacts caused by low-rank reconstruction, such as vertical and horizontal streaks in images, are also reduced (see Fig. \ref{fig:visual_results}(c)). Each patch is encoded into its latent components, and they are individually approximated before the data is spatially reassembled. Further processing (e.g. "smoothing" convolutions) is applied to this reassembled output.
\vspace{-5pt}
\section{Experimental Validation}
\vspace{-5pt}
For our experiments, we look at (1) A basic encoder/decoder VQVAE adapted from Google Deepmind 
(2) Our ViT-inspired transformer encoder that acts on patched data, feeding into a low-rank reconstruction, then an optional lightweight smoothing convolution pair. Each model is trained over 50 epochs with batch sizes of 128-512, learning rate = 3e-4, and weight decay = 1e-5, on 1-3 NVIDIA 40GB A100 GPU(s) depending on memory requirement. We use the Adam optimizer and choose mean squared error for loss. 


\begin{figure}
  \centering
  \includegraphics[width=0.9\linewidth]{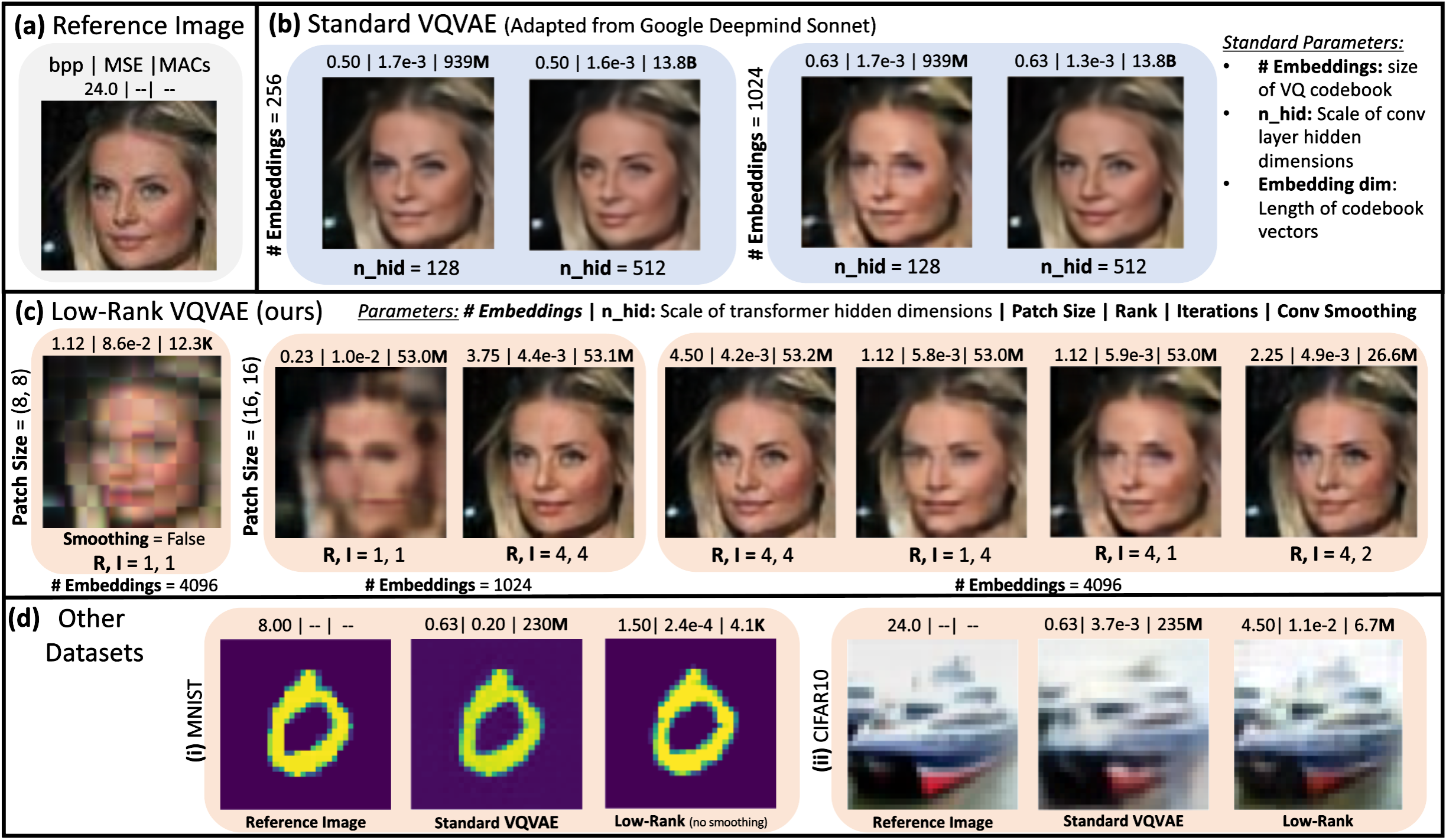}
  \caption{Visual comparison of various parameters across standard VQVAEs (b) and our proposed low-rank VQVAE (c). We adjust various parameters and observe both their visual quality, as well as bits-per-pixel (bpp), MSE, and computational overhead in multiply and accumulates (MACs). Datasets used are CelebA 64x64, CIFAR-10, and MNIST \cite{celeba, cifar10, mnist}.}
  \label{fig:visual_results}
\end{figure}

\begin{figure}
  \centering
  \includegraphics[width=0.65\linewidth]{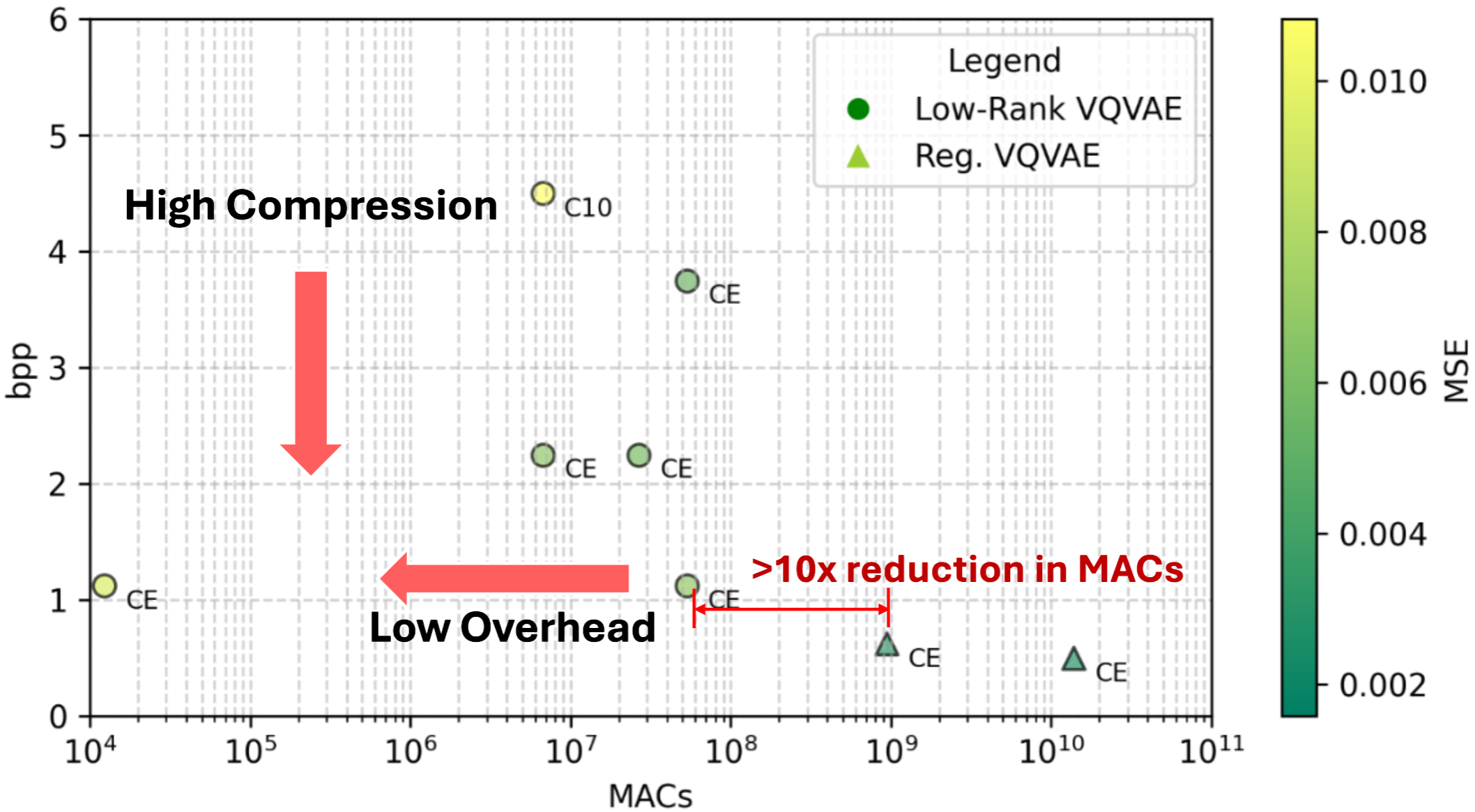}
  \caption{Graph comparison of various models' compression (bpp) and their respective decoder overhead as measured in MACs. Reconstruction quality is reported in color scale as MSE.}
  \label{fig:scatter_plot}
  \vspace{-10pt}
\end{figure}

\textbf{Pursuing High-Fidelity Low-Overhead Reconstruction:}
The quality of our framework comes from: (a) the quantity of low-rank components that kick off the decoding process, and (b) the strength of the smoothing decoder. Higher rank and iterations lead to significant increases in bpp, though, so they must be limited. Furthermore, when patching the image to gain finer detail, the number of low-rank components necessary to reconstruct the image increases. These relationships can be seen in Fig. \ref{fig:visual_results}(c). The majority of our overhead still comes from decoder convolutions if they are present; however, since the images are partially reconstructed by low-rank-approximated patches, the necessary compute has been drastically reduced. With upwards of two orders of magnitude fewer operations, we can achieve comparable quality, albeit at slightly higher bpp (Fig. \ref{fig:scatter_plot}).



\section{Conclusion}


In this work, we developed an alternative formulation of VQ-VAEs, with the major innovation in simplifying the computation of the decoder block. By introducing efficient low-rank computations, high-fidelity output can be approximated in a single step based on the vector-quantized latent representation. This approach reduces the computational cost of reconstruction by orders of magnitude while still allowing for compression and further model-based denoising. Looking ahead, the framework could be extended to high-resolution data and generative tasks, where it may significantly decrease the number of steps required to produce high-quality outputs.



\printbibliography

\end{document}